\documentclass{article}

\usepackage[accepted,nohyperref]{icml2021}
\usepackage[ref]{leaf}

\newcommand{\uvec}{\mbf{u}}  
\newcommand{\Uvec}{\mbf{U}}  

\newcommand{\codeurl}{\href{https://u.perhapsbay.es/simplex-gp-code}{\texttt{u.perhapsbay.es/simplex-gp-code}}}

\icmltitlerunning{Kernel Interpolation on the Permutohedral Lattice for Scalable Gaussian Processes (Simplex-GPs)}

\begin{document}

\twocolumn[
\icmltitle{SKIing on Simplices: Kernel Interpolation on the Permutohedral Lattice for Scalable Gaussian Processes}

\icmlsetsymbol{equal}{*}

\begin{icmlauthorlist}
\icmlauthor{Sanyam Kapoor}{equal,nyu}
\icmlauthor{Marc Finzi}{equal,nyu}
\icmlauthor{Ke Alexander Wang}{stanford}
\icmlauthor{Andrew Gordon Wilson}{nyu}
\end{icmlauthorlist}

\icmlaffiliation{nyu}{New York University, NY, USA}
\icmlaffiliation{stanford}{Stanford University, CA, USA}

\icmlcorrespondingauthor{Sanyam Kapoor}{sanyam@nyu.edu}

\icmlkeywords{Gaussian processes, permutohedral lattice}

\vskip 0.3in
]

\printAffiliationsAndNotice{\icmlEqualContribution}

\begin{abstract}
State-of-the-art methods for scalable Gaussian processes use 
iterative algorithms, requiring fast matrix vector multiplies (MVMs) 
with the covariance kernel. The Structured Kernel Interpolation (SKI) 
framework accelerates these MVMs by performing efficient MVMs on a grid 
and interpolating back to the original space. In this work, we develop 
a connection between SKI and the permutohedral lattice used for 
high-dimensional fast bilateral filtering. Using a sparse simplicial grid 
instead of a dense rectangular one, we can perform GP inference 
exponentially faster in the dimension than SKI. Our approach, 
Simplex-GP, enables scaling SKI to high dimensions, while maintaining 
strong predictive performance. We additionally provide a CUDA 
implementation of Simplex-GP, which enables significant GPU 
acceleration of MVM based inference.
\end{abstract}

\section{Introduction}
\label{sec:intro}
Gaussian processes (GPs) are widely used where uncertainty is critical to the task at hand 
\citep{Deisenroth2013ASO,Frazier2018ATO,Balandat2019BoTorchPB}.
At the same time, datasets in machine learning applications are growing 
in not only size but also dimensionality \citep{deng2009imagenet,
brown2020language}. To address the cubic complexity of exact inference 
with GPs, past works have proposed a myriad of approximation methods 
\citep{Snelson2007LocalAG,Titsias2009VariationalLO,hensman2013gaussian,
Hensman2015ScalableVG,Hensman2015MCMCFV,Salimbeni2017DoublySV,
Izmailov2018ScalableGP,Gardner2018GPyTorchBM,Liu2020WhenGP}. 
\emph{Structured Kernel Interpolation} (SKI), proposed by 
\citet{Wilson2015KernelIF}, is particularly effective on large datasets
\citep{Wilson2015ThoughtsOM}.

\begin{figure}[!t]
\centering
\includegraphics[width=.8\linewidth]{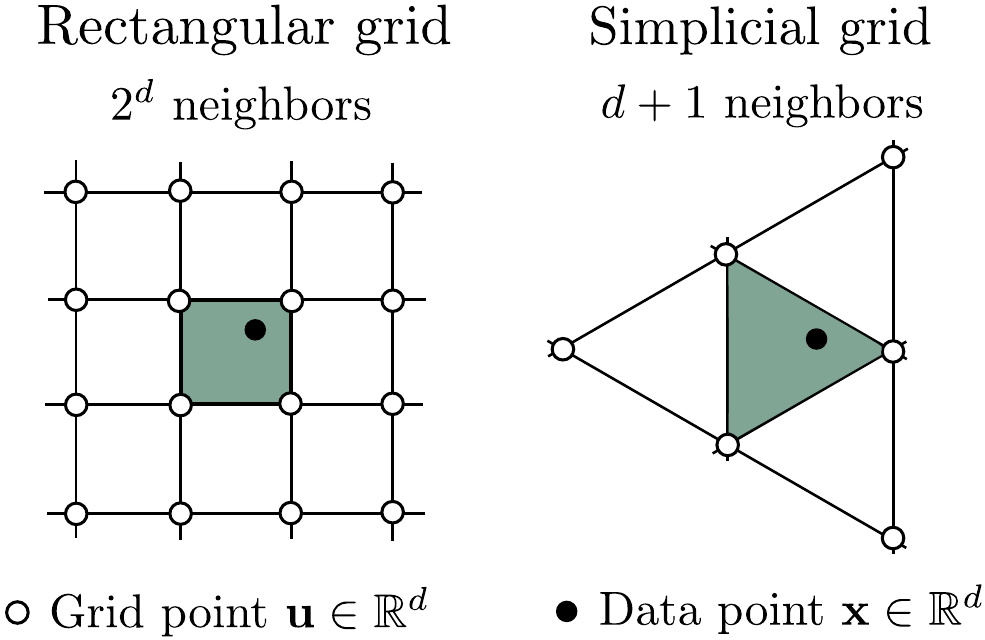}
\caption{Comparison of the number of grid points required in SKI 
\citep{Wilson2015KernelIF} against our method. SKI requires 
exponentially more number of grid points due to its dense lattice 
construction. Simplex-GP on the other hand can get away with only a 
sparse set of points on the permutohedral lattice.}
\label{fig:neighbors}
\end{figure}

SKI tiles the data space with a dense rectangular grid of inducing points.
By interpolating kernels to the grid, and exploiting structure in the matrices,
SKI greatly reduces the computational cost of inference 
to be only linear in the dataset size and sub-quadratic in the number 
of grid points. The powerful grid structure of SKI, however, is also 
its downfall in high dimensional settings, since SKI requires at least
$2^d$ grid points for a $d$-dimensional dataset. Thus, SKI 
requires time and memory exponential in the dataset dimensionality, 
making it unsuitable for datasets with more than $\sim 5$ dimensions. 
Building on SKI, \citet{Gardner2018ProductKI} proposed SKIP, which reuses a one-dimensional grid across all 
dimensions and builds up low rank approximations to the covariance matrices, reducing inference time to be linear in $n$ as well as $d$. However, even with these improvements, SKIP is limited to $d\sim 10$ to $30$ for 
large datasets since the method uses memory equivalent to storing 
$\sim 20\cdot d$ copies of the training dataset, and the low rank approximation can sometimes be limiting.

Dimensionality is not a challenge in GP inference alone. In image 
filtering, a broad class of non-linear edge-preserving filters including the bilateral 
filter, non-local means, and other related filters have found growing 
importance \citep{Aurich1995NonLinearGF,Tomasi1998BilateralFF,
Paris2006AFA}. Such filters map a set of vector values
${\{\yvec_i\} \subseteq \reals^n}$  (typically the pixel values in an image) to 
${\{\yvec^\prime_i\} \subseteq \reals^n}$ where each $\yvec^\prime_i$ 
is a Gaussian weighted linear combination of vectors at locations 
${\{\xvec_i\}_{i=1}^n \subseteq \reals^d}$:
\begin{align}
\yvec_i^\prime = \sum_{j=1}^n e^{- \norm{\xvec_j - \xvec_i}^2 } \yvec_j ~,\label{eq:nl_filters}
\end{align}

which can be rewritten as a matrix-vector product involving a matrix 
with entries ${(K_{\Xvec, \Xvec})_{ij} = 
e^{- \norm{\xvec_j - \xvec_i}^2 }}$.

The naive implementation is computationally expensive to be run on images,
and to remedy this, many fast approximations have been developed
including the popular permotohedral lattice \citep{Adams2010FastHF},
enabling real time filtering rates on images and in fairly high dimensions.

In these seemingly disparate fields of GP inference and 
high-dimensional Gaussian filtering, we find complementary strengths. 
On one hand, GP inference is made scalable by structure-exploiting 
algebra through the SKI framework, but is limited by the exponential 
scaling in dimension. On the other hand, high-dimensional Gaussian 
filtering relies on a similar sparse linear combination of the 
neighborhood, and makes scaling with respect to dimension favorable by 
embedding points onto the permutohedral lattice. Our work combines 
these strengths into an inference algorithm for Gaussian processes by 
computing \cref{eq:nl_filters} over the permutohedral lattice using 
matrix-vector multiplications.

Inspired by the connection between image filtering and kernel methods, 
we develop Simplex-GP, an interpolation-based scalable Gaussian process 
approximation method that leverages advances in high-dimensional image 
filtering. Instead of using a dense cubic grid as in SKI, we employ a sparse
simplicial lattice. In the SKI framework, observations must be interpolated to their neighbors in the lattice.
Crucially, while each data point in the cubic lattice has $2^d$ neighbors, each point in a simplicial lattice has only $d+1$ neighbors, as illustrated in \cref{fig:neighbors}. 
This fact allows us to use exponentially fewer inducing points than in SKI, and perform GP inference in time $\bigo{d^2(n + m)}$ and $\bigo{dm}$ memory
where $m$ is the number of inducing points.


Our key 
contributions are summarized below:

\begin{itemize}
\item Drawing parallels between image filtering and Gaussian process 
inference, we develop a novel kernel interpolation scheme using the 
permutohedral lattice \citep{Adams2010FastHF}. Our method can perform 
GP inference in $\bigo{n\cdot d^2}$ time, exponentially faster than SKI 
which requires $\bigo{n\cdot 2^d}$ time, for a set of $n$ 
$d$-dimensional inputs.

\item To allow optimization of the marginal likelihood with respect to 
kernel hyperparameters using off-the-shelf automatic 
differentiation-based optimizers \citep{Paszke2019PyTorchAI}, we show 
how gradients can also be efficiently formulated as filtering 
operations on the lattice alone.

\item We extend the permutohedral lattice to include filtering with 
respect to general stationary kernels such as the Mat\'ern kernel 
\citep{rasmussen2005gaussian}.

\item We provide efficient CPU and CUDA accelerated implementations of 
permutohedral lattice filtering at \codeurl. Our
implementations are compatible with GPyTorch \citep{Gardner2018GPyTorchBM}, 
making them amenable for easy use with other numerical methods.
\end{itemize}

In summary, our method Simplex-GP provides a scalable extension to the 
SKI framework, which alleviates both the low-dimensional limitations of 
KISS-GP \citep{Wilson2015KernelIF}, and high-memory requirements of 
SKIP \citep{Gardner2018ProductKI}, by instead packing the inputs into a 
more efficient lattice. This is followed by significant performance 
gains over SKIP, reducing the performance gap to exact 
Gaussian processes.

\section{Gaussian Processes}
\label{sec:gp}

By expressing priors over functions, GPs allow us to build
flexible non-parametric function approximators that can learn structure
in data through covariance functions \citep{rasmussen2005gaussian}. In 
a typical regression setting, for a dataset $\dset$ of size $n$, we 
model the relationship between inputs (predictors) 
${\Xvec = \left[ \xvec_1,\xvec_2,\dots,\xvec_n \right] 
\in \reals^{n \times d}}$ and corresponding outputs 
${\yvec = \left[ y(\xvec_1),y(\xvec_2),\dots,y(\xvec_n) \right]^\top 
\in \reals^{n}}$ using a Gaussian process prior, 
${\fvec = \left[ f(\xvec_1),f(\xvec_2),\dots,f(\xvec_n) \right]^\top 
\sim \gp{\mbf{\mu}_{\Xvec}, K_{\Xvec,\Xvec}}}$, determined by a mean 
vector ${(\mbf{\mu}_{\Xvec})_i = \mbf{\mu}(\xvec_i)}$ and a covariance 
matrix ${(K_{\Xvec,\Xvec})_{ij} = k_\theta(\xvec_i, \xvec_j)}$. The 
kernel function $k_\theta$ is parametrized by $\theta$. The observation 
likelihood is assumed to be input-independent Gaussian additive noise, 
i.e. ${y(\xvec) \mid f(\xvec) \sim \gaussian{f(\xvec), \sigma^2}}$ with 
variance $\sigma^2$.

In maximum likelihood inference, we compute the posterior over functions ${p(f \mid \dset, \theta, \sigma^2)}$ using parameters 
${\{\theta,\sigma^2\}}$ that maximize the marginal $\log$-likelihood 
${\log{p(\yvec\mid\Xvec)}}$. For the chosen prior and likelihood pair, 
the predictive distribution at $n_\star$ test inputs is a multivariate 
Gaussian $\gaussian{\mean{\mbf{f}_\star},\cov{\mbf{f}_\star}}$ where,
\begin{align}
\mean{\mbf{f}_\star} =&~\mbf{\mu}_{\Xvec_\star} +  K_{\Xvec_\star,\Xvec}\left[K_{\Xvec,\Xvec} + \sigma^2\mbf{I} \right]^{-1} \yvec, \label{eq:gp_pred_mean} \\
\begin{split}
\cov{\mbf{f}_\star} =&~K_{\Xvec_\star,\Xvec_\star} -\\ &~ K_{\Xvec_\star,\Xvec}\left[K_{\Xvec,\Xvec} + \sigma^2\mbf{I} \right]^{-1} K_{\Xvec,\Xvec_\star}.
\end{split} \label{eq:gp_pred_cov}
\end{align}

The marginal $\log$-likelihood (MLL) is given by,
\begin{align}
\begin{split}
\log{p(\yvec\mid\Xvec)} \propto &- \tfrac{1}{2} \yvec^\top\left( K_{\Xvec,\Xvec} + \sigma^2\mbf{I} \right)^{-1}\yvec \\ &-\tfrac{1}{2}\logdet{K_{\Xvec,\Xvec} + \sigma^2\mbf{I}}.    
\end{split}
 \label{eq:gp_mll}
\end{align}

Owing to inverse and determinant computations in 
\cref{eq:gp_pred_mean,eq:gp_pred_cov,eq:gp_mll}, naive inference takes 
$\bigo{n^3}$ compute and $\bigo{n^2}$ storage, which is prohibitively 
expensive for large $n$. Modern state-of-the-art and scalable 
implementations of GP regression \citep{Gardner2018GPyTorchBM,
Matthews2017GPflowAG}, however, rely on \emph{Krylov subspace methods} 
like \emph{conjugate gradients} (CG) \citep{Golub1996MatrixC}; more 
recently, \citet{Wang2019ExactGP} demonstrated that exact Gaussian 
processes can be scaled to more than a million input data using such 
methods. These methods depend only on \textit{Matrix Vector 
Multiplications} (MVMs): ${\mathbf v \mapsto 
K_{\Xvec,\Xvec} \mathbf v}$ rather than computing the elements of 
$K_{\Xvec,\Xvec}$ itself. The complexity of one inference step is 
reduced to $\bigo{pn^2}$ for $p$ CG iterations, where typically 
${p \ll n}$ suffices in practice.

Alternatively, inducing point methods \citep{Candela2005AUV,
Titsias2009VariationalLO,hensman2013gaussian,Hensman2015ScalableVG} 
have been introduced as a scalable approximation to exact GPs. A set of 
$m$ inducing points or \emph{pseudo}-inputs are introduced as 
parameters, with usual GP inference built on top of these points. 
One-step inference costs only $\bigo{m^2n + m^3}$ compute and 
$\bigo{m^2 + mn}$ storage, but the cross-covariance term corresponding 
to the $m^2n$ factor can still be costly, limiting $m \ll n$.

\subsection{Structured Kernel Interpolation}

\citet{Wilson2015KernelIF} show that using \emph{Structured Kernel 
Interpolation} (SKI) to introduce algebraic structure, we can use 
extremely high number of inducing points, irrespective of the nature of 
the input space. Using simple sparse interpolation schemes (e.g. 
inverse distance weighting or spline interpolation), one can accelerate 
the computations required for computing the covariance matrices via 
KISS-GP \citep{Wilson2015KernelIF,Wilson2015ThoughtsOM}.

Given a set of $m$ inducing points ${\Uvec = 
[\uvec_1,\uvec_2,\dots,\uvec_m]}$, \citet{Wilson2015KernelIF} propose a 
general approximation to the cross-covariance kernel matrix 
${K_{\Xvec,\Uvec} \in \reals^{n\times m}}$ as 
${\widetilde{K}_{\Xvec,\Uvec} = W_\Xvec K_{\Uvec,\Uvec}}$, where 
${W_\Xvec \in \reals^{n\times m}}$ describe the interpolation weights. 
Subsequently, one can plug this into the subset of regressors (SoR) 
\citep{Silverman1985SomeAO,Candela2005AUV} formulation, and prior 
covariance is approximated by \cref{eq:k_ski}.
\begin{align}
K_{\Xvec,\Xvec} \approx  W_\Xvec K_{\Uvec,\Uvec} W_\Xvec^\top \defeq \widetilde{K}_{\Xvec,\Xvec}~. \label{eq:k_ski}
\end{align}

The SKI framework now allows one to posit efficient structures for the 
inducing points irrespective of the nature of the input points, such 
that the matrix $K_{\Uvec,\Uvec}$ inherits efficient computation. For 
instance, placing the inducing points on a rectilinear grid allows one 
to exploit Toeplitz or Kronecker algebra \citep{Wilson2015KernelIF}. 
Further, by using \emph{local} interpolations (e.g. inverse distance 
weighting, or spline interpolation in \citet{Wilson2015KernelIF}) 
instead of \emph{global} GP interpolations, the sparsity induced in the 
weights matrix $W_\Xvec$ greatly improves scalability by allowing 
$m \gg n$. The inference proceeds by plugging 
${\widetilde{K}_{\Xvec,\Xvec} + \sigma^2\mbf{I}}$ into a conjugate 
gradients and an eigenvalue solver for $\log$-determinant computations. 
This method, termed as KISS-GP, is also amenable to general 
\emph{Krylov subspace methods} \citep{Golub1996MatrixC,
Gardner2018GPyTorchBM}.

When exploiting Toeplitz structure in the inducing points, KISS-GP 
requires $\bigo{n2^d + m\log{m}}$ compute and $\bigo{n+m}$ storage. 
When exploiting Kronecker structure in the inducing points, KISS-GP 
requires $\bigo{n2^d + dm^{1+1/d}}$ compute and $\bigo{n + dm^{2/d}}$ 
storage. The number of inducing points $m$, however, grow exponentially 
in number with the dimension, as a consequence of being embedded in a 
\emph{rectangular} grid, limiting the application to small dimensions. 
Within the SKI framework, \citet{Gardner2018ProductKI} instead propose 
SKIP, which provides an efficient application of MVMs to stationary 
kernels that can be decomposed as Hadamard products across dimensions. 
Low rank approximations in SKIP, however, can reduce performance.

Our work lies within the SKI framework, and provides a scalable 
approach which alleviates both the low-dimensional limitations of 
KISS-GP \citep{Wilson2015ThoughtsOM}, and high-memory requirements of 
SKIP \citep{Gardner2018ProductKI}, by instead packing the inputs into a 
more efficient lattice, while performing significantly better on our 
benchmark datasets.

\section{Bilateral Filtering}
\label{sec:filtering}

Gaussian processes aside, bilateral filtering is a popular method in 
the image processing community \citep{Aurich1995NonLinearGF,
Tomasi1998BilateralFF}. The Gaussian blur filter is a common tool in 
signal processing, computer vision, computational photography, and 
elsewhere for smoothing out noise, irregularities, and unwanted high 
frequency information. In the image domain especially, signals have 
hard discontinuities such as those caused by occlusion. Special 
\emph{edge-preserving} filters have been developed to smooth signals 
while preserving meaningful edges and boundaries, the most popular of 
which is the bilateral filter \citep{Paris2006AFA}.

Extending the Gaussian blur, the bilateral filter is a nonlinear filter 
that is Gaussian in both spatial distance and the RGB difference in 
intensity. Mathematically, the blurred pixels have the value
\begin{equation}
    \mbf{v}'_i = \sum_i\left(e^{-\norm{\mbf{p}_i-\mbf{p}_j}^2/2\sigma_p^2 - \norm{\mbf{v}_i-\mbf{v}_j}^2/2\sigma_v^2}
    \right)\mbf{v}_j~,
\end{equation}
where $\mbf{v}_i\in \reals^3$ is a vector of the RGB pixel intensities 
for a given pixel index $i$ and $\mbf{p}_i \in \ints^2$ is the 
horizontal and vertical coordinates of the pixel on the grid.

The bilateral filter has proven effective in many low level image tasks 
such as haze removal \citep{zhu2015fast}, contrast adjustment, detail 
enhancement \citep{fattal2007multiscale}, image matting, depth 
upsampling \citep{barron2016fast}, and even semantic segmentation 
\citep{Krhenbhl2011EfficientII}. A variant of the bilateral filter 
known as the joint bilateral filter allows for blurring of an image 
$\mbf{v}$ while respecting the edges in a second reference image 
$\uvec$, and in general concatenating 
${\xvec :=\mathrm{Concat}(\mbf{p}/\sigma_p,\uvec/\sigma_u)}$ we can 
write the operation more abstractly as \cref{eq:nl_filters}.

A major challenge with bilateral filtering is the $\bigo{n^2}$ 
computation cost which is extremely costly for images. Various 
approaches have been proposed in the literature to accelerate the 
computation such as the Fast Gauss Transform \citep{greengard1991fast}, 
Gaussian KD-tree \citep{adams2009gaussian}, Guided Filter 
\citep{he2010guided}, and the permutohedral lattice 
\citep{Adams2010FastHF}. Due its favorable $O(nd^2)$ scaling and the 
generality of the method, the permutohedral lattice has been the method 
of choice for downstream applications \citep{Krhenbhl2011EfficientII,
kiefel2014permutohedral,su2018splatnet}.

\subsection{GP Inference with the RBF Kernel is Equivalent to Bilateral 
Filtering}
\label{sec:gp_is_bf}

Efficient inference in Gaussian processes depends directly on having an 
efficient way to perform matrix vector multiplications using the kernel 
matrix $K_{\Xvec,\Xvec}$. For RBF kernels, after normalizing by 
lengthscale $x_i \mapsto x_i/\ell$, this matrix is precisely the same 
as the one that defines bilateral filtering,
${k(\xvec_i,\xvec_j)=e^{-\tfrac{1}{2}\norm{\xvec_j-\xvec_i}^2}}$. In 
both cases, these vectors can lie off the grid because of the color 
intensity in bilateral filter and naturally ungridded data for GPs. In 
this general framing, methods that are used to accelerate image 
filtering, can also be used to accelerate GP inference (and vice-
versa).

\begin{figure}[!ht]
\centering
\includegraphics[width=\linewidth]{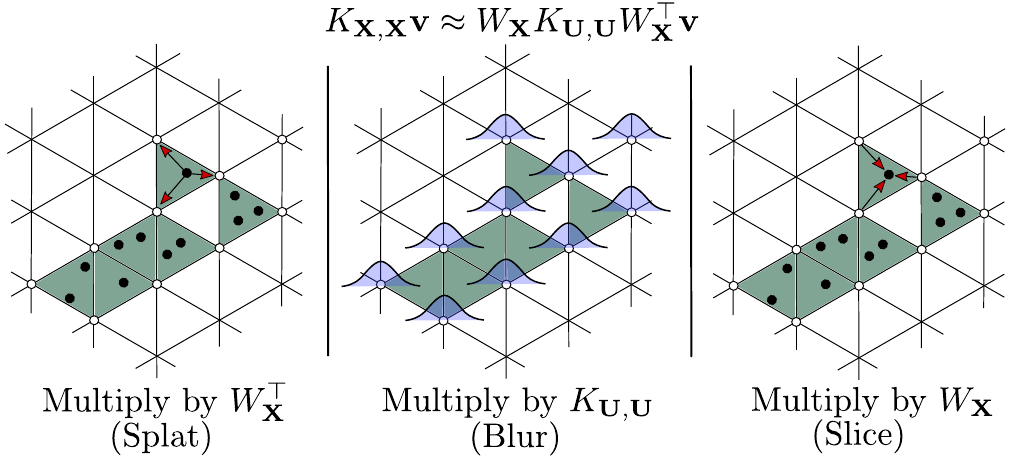}
\caption{Filtering using the permutohedral lattice involves three 
stages - \emph{Splat}, \emph{Blur}, and \emph{Slice}. The usual 
convolutions required for \emph{Blur} are wrapped by a projection onto 
the lattice via barycentric interpolation (\emph{Splat}), and finally 
resampling the lattice at input locations for the filtering output 
(\emph{Slice}). This is equivalent to the SKI decomposition, but with a 
different grid.}
\label{fig:correspondence}
\end{figure}

The SKI and the permutohedral lattice filtering, despite arising from 
separate communities, are in fact closely related. In the SKI 
decomposition,
${K_{\Xvec,\Xvec} \approx  W_\Xvec K_{\Uvec,\Uvec} W_\Xvec^\top}$ the
linear operations $\mbf{v} \mapsto W_\Xvec^\top \mbf{v}$, $\mbf{w} \mapsto K_{\Uvec,\Uvec} \mbf{w}$ and 
$\mbf{v}^\prime \mapsto W_\Xvec \mbf{v}^\prime$ map \emph{precisely} to the \emph{Splat}, 
\emph{Blur}, and \emph{Slice} stages respectively of filtering with the 
permutohedral lattice \citep{Adams2010FastHF}. \emph{Splat} 
interpolates the values onto the lattice points, \emph{Blur} applies 
the filter on the lattice, and \emph{Slice} interpolates back to the 
original locations. SKI uses linear interpolation to a cubic lattice, 
whereas the permutohedral filtering uses barycentric interpolation to a 
simplicial lattice. While SKI stores values densely in the lattice with 
dense arrays, the permutohedral lattice stores values sparsely using a 
hash table. With SKI, the cubic lattice ${K_{\Uvec,\Uvec} = 
K_{\Uvec_1,\Uvec_1} \otimes \cdots \otimes K_{\Uvec_d,\Uvec_d}}$ 
induces Kronecker structure to enable efficient computation, whereas 
with the permutohedral lattice the computation splits over the (non-
orthogonal) lattice directions. We visualize this correspondence in 
\cref{fig:correspondence}, and review permutohedral lattice filtering 
in the context of kernel interpolation.

\subsection{Permutohedral Lattice}
\label{sec:pl}

Grids have been used in the past not just for scalable GPs but also 
bilateral filtering \citep{chen2007real}. An important distinction from 
prior approaches, \citet{Adams2010FastHF} discovered that the key to 
avoiding the curse of dimensionality is in using a simplicial lattice 
rather than a cubic one, which they term a permutohedral lattice.

The \emph{permutohedral lattice} is a higher-dimensional generalization of 
the hexagonal lattice built from triangles in $\reals^2$ to simplices in 
$\reals^d$. Each of these simplicies are the same, and can best understood 
as permutations of a \emph{canonical simplex}, with vertices 
${\mbf{s}_0,\mbf{s}_1,\dots,\mbf{s}_d}$ given by,
\begin{equation}
\mbf{s}_k = [~\underbrace{k,\dots,k}_{d + 1 - k}, \underbrace{k - (d + 1),\dots,k - (d + 1)}_{k}~]~. \label{eq:canon_simplex}    
\end{equation}

These vertices form a basis of a $d$ dimensional hyperplane $H_d$. See \citet{Baek2012LatticeBasedHG} 
for a detailed exposition on the theoretical properties of the 
permutohedral lattice. Following the terminology in 
\citet{Adams2010FastHF}, filtering using the permutohedral lattice involves 
three operations -- \emph{Splat}, \emph{Blur}, and \emph{Slice}. In 
essence, the filtering is carried out by the \emph{Blur} operation. The 
\emph{Splat} and \emph{Slice} operations provide the mapping of the 
position vectors, between the input space and the lattice embedding space. 
\cref{fig:correspondence} illustrates operations on the permutohedral 
lattice for two-dimensional inputs.

\paragraph{Splat} The input vectors $\xvec$ are first embedded into the 
subspace defined by $H_d$. While the linear map can be constructed directly 
using the basis defined by \cref{eq:canon_simplex}, one can more 
efficiently compute the embedding in linear time, i.e. $\bigo{d}$, by 
instead using a more efficient triangular basis, which we denote by 
$\mbf{E}$. 

Surrounding the embedded point, one can find the enclosing simplex and its 
vertices by a rounding algorithm and comparison to a canonical simplex in a 
total $\bigo{d^2}$ time \citep{Conway1988SpherePL}. These vertices serve as 
the inducing points for Simplex-GP to which the value $v$ is projected.

Given the bounding vertices $\mbf{s}_k$ of the given point inside the simplex 
(of which there are only $d+1$ rather than $2^d$ for a cubic lattice), 
barycentric interpolation weights $w_k$ are computed based on the proximity 
to each of the vertices $\norm{\xvec - \mbf{s}_k}$. These weights are the $d+1$ 
nonzero values in the matrix in each row $W_\xvec$ of $W_\mbf{X}$, giving it the 
sparse structure similar to the interpolation matrix in KISS-GP. These 
inducing point vertices are stored in a hash table along with the splatted 
values $W_\mbf{X}^\top \mbf{v}$. Unlike KISS-GP, the lattice vertices (inducing 
points) which do not border any point $\xvec$ are not computed or stored.

The amortized hash-table lookup compute is then $\bigo{d}$. For a total of 
$m$ hashtable entries corresponding to the generated lattice points (inducing points), the 
total storage needed is $\bigo{md}$, where $m$ itself is loosely upper-
bounded by $\bigo{nd}$. We empirically demonstrate the significant storage 
gains due to embedding into the permutohedral lattice. Each input $\xvec$ 
is now represented as a barycentric interpolation of the lattice points 
$\uvec$ in the enclosing simplex.

\paragraph{Blur}  Having constructed the lattice, the \emph{Blur} operation 
is simply a convolution with the neighbors in the lattice using an 
appropriately-sized stencil across each direction in the lattice 
sequentially ($d+1$ in total). There are $\bigo{d}$ neighbors to lookup, 
and each lookup in the hashtable requires $\bigo{d}$ time. Therefore, each 
blur step can be completed in $\bigo{d^2}$ time. As an example, a Gaussian 
blur can be implemented by convolving with the binomial stencil 
${\left[.5,1,.5\right]}$ across each direction in the lattice. In 
\cref{sec:disc_kernels}, we describe a general approach to build stencils 
for any stationary kernel in Gaussian processes. We note however that in 
principle lattice points could be added to the hash table in the blurring 
along one axis that could affect the subsequent blur steps along the other 
directions; however, like \citep{Adams2010FastHF} we ignore this second 
order effect and keep the number of inducing points fixed during the 
filtering operation.

\paragraph{Slice} The \emph{Slice} operation resamples the values back at 
the input location using barycentric interpolation weights as before (the 
transpose of the sparse matrix $W_X^\top$). By caching the barycentric 
weights for each generated lattice point during splatting, we can avoid any 
redundant computations, and recover the values in $\bigo{d}$ time. A 
complete scan over all the inputs will therefore take $\bigo{nd}$.

In summary, the entire algorithm for filtering using the permutohedral 
lattice requires $\bigo{d^2(n + m)}$ compute, and $\bigo{dm}$ storage.

\section{Simplex Gaussian Processes}
\label{sec:simplex_gp}

We have established in \cref{sec:gp_is_bf} that bilateral filtering, and 
RBF kernel MVMs in GP inference are fundamentally the same. \cref{sec:pl} 
describes an accelerated approach to high-dimensional Gaussian filtering. 
Consequently, our proposed inference method, named Simplex Gaussian 
processes (Simplex-GPs), effectively leverage this connection by replacing 
the exact MVMs with accelerated bilateral filtering using the permutohedral 
lattice. In summary, the SKI decomposition \cref{eq:k_ski}, and the 
corresponding bilateral filtering stages are annotated in 
\cref{eq:ski_to_bf}. The computational complexity of inference in 
Simplex-GPs is listed alongside other GP inference methods in 
\cref{table:complexities}.
\begin{align}
K_{\Xvec,\Xvec} \approx \underbrace{W_{\Xvec}}_{\mathrm{Slice}}\underbrace{K_{\Uvec,\Uvec}}_{\mathrm{Blur}}  \underbrace{W_{\Xvec}^\top}_{\mathrm{Splat}}~.    \label{eq:ski_to_bf}
\end{align}
Notably, in this decomposition, only the blur matrix $K_{\Uvec,\Uvec}$ 
depends on the choice of the stationary kernel used to model the Gaussian 
process prior. The \emph{Splat} $W_{\Xvec}^\top$ and \emph{Slice} 
$W_{\Xvec}$ matrices remain the same.

Without further changes, the bilateral filtering on the permutohedral 
lattice is only equivalent to a MVM with RBF kernel. However, we may wish 
to use other stationary kernels with GP inference. Stationary kernels that 
depend only on the difference between the points, 
${\tau = \lvert \uvec_i - \uvec_j\rvert}$, which includes a broad class of 
kernels, including the RBF kernel and the Mat\'ern kernel
\citep{rasmussen2005gaussian}. To adapt our method to another stationary 
kernel like, we simply have to compute an appropriate stencil of 
coefficients to build a discrete approximation of the non-Gaussian kernel 
$k(\tau)$. We now develop this approximation.

\begin{table}[!ht]
\caption{Time complexities for MVM used during Gaussian process inference. 
$n$ is the number of inputs, $m$ is the number of inducing points, and $r$ is the
rank used in the SKIP approximation (typically between $20$ and $100$).}
\label{table:complexities}
\vskip 0.15in
\begin{adjustbox}{width=\linewidth}
\begin{sc}
\begin{tabular}{l|c}
\toprule
Method & Time Complexity of one MVM \\
\midrule
Exact & $\bigo{n^2}$ \\
KISS-GP (Kronecker) & $\bigo{n2^d}$ \\
SKIP & $\bigo{rnd}$ \\
Simplex-GP & $\bigo{nd^2}$ \\
\bottomrule
\end{tabular}
\end{sc}
\end{adjustbox}
\end{table}

\subsection{Discretizing Generic Stationary Kernels}
\label{sec:disc_kernels}

We can compute the discretized stencil coefficients by evaluating the 
kernel along the lattice points. Let \(\xvec'\) be \(i\) points apart from 
\(\xvec\) along a given lattice direction \(\mbf s_k\) such that 
${\xvec'=\xvec+i\mbf s_k}$. For a stationary kernel $k$, we then have 
$k(\xvec, \xvec') = k(\norm{\xvec-\xvec'}^2)=k(is)$ where 
${s := \norm{\mbf{s}_k}}$ is the size of the lattice spacing. We can 
approximate \(k(\xvec, \cdot)\) by only considering the points \(\xvec'\) 
that lie within $i$ spaces apart with $i=-r,...,r$. We can then choose $r$ 
based on how quickly the tails of $k$ decay in order to reach a desired 
error threshold. Even with a fixed number of evenly spaced points 
at which to evaluate the kernel, there is an additional free parameter $s$, 
corresponding to the spacing between points, which we must determine in 
order to maximize the accuracy of our stencil for $k$.

\begin{figure}[!ht]
\centering
\includegraphics[width=\linewidth]{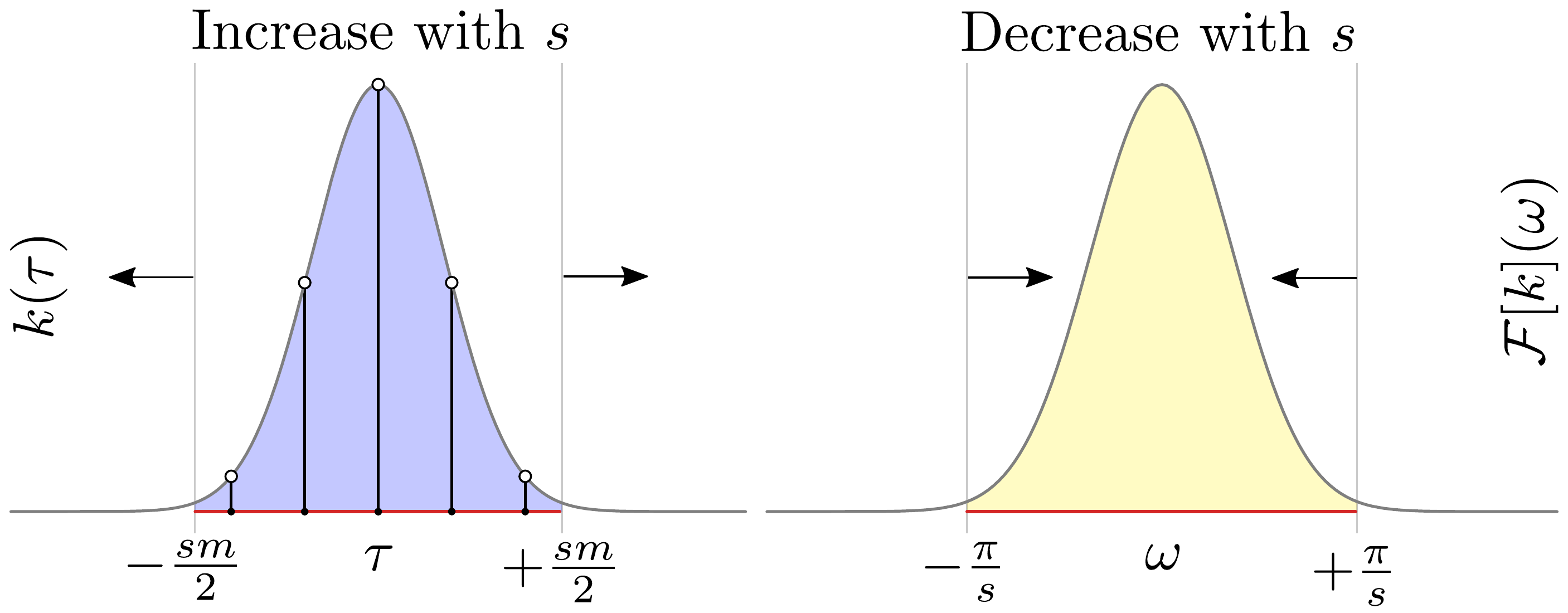}
\caption{Increasing the coverage in the spatial domain $\tau$ (left), 
reduces the coverage in Fourier domain $\omega$ (right), monotonically due 
to aliasing. Finding a good spacing $s$ for the discretization amounts to 
balancing the coverage defined by \cref{eq:balance_coverage}. Each vertical 
stick in the spatial domain covers a region $[-s/2,s/2]$ around itself.}
\end{figure}

When choosing this spacing, we must balance the coverage of the underlying 
kernel function in both the spatial and frequency domains 
\citep{Birchfield}. For a fixed number of points, increasing the spacing 
will cover a larger fraction of the kernel function in the spatial domain, 
but a reduce coverage of the function in the Fourier domain due to 
aliasing. Given $m=2r+1$ discretization points with \(r\geq 0\), our 
stencil in the spatial domain covers the interval $[-sm/2,sm/2]$. Since the 
points are spaced $s$ apart, the Nyquist frequency is 
${f_{\text{ny}} = 1/(2s)}$ or 
${\omega_{\text{ny}}=2\pi f_{\text{ny}}=\pi/s}$ in radians. Therefore, 
matching coverage of $k$ in both domains requires that 
\begin{equation}\label{eq:balance_coverage}
    \frac{\int_{-sm/2}^{sm/2}k(\tau)d\tau}{\int_{-\infty}^{\infty}k(\tau)d\tau} = \frac{\int_{-\pi/s}^{\pi/s}\mathcal{F}[k](\omega)d\omega}{\int_{-\infty}^{\infty}\mathcal{F}[k](\omega)d\omega}~. 
\end{equation}

Since both $k(\tau)$ and $\mathcal{F}[k](\omega)$ are strictly positive, 
the left-hand-side increases monotonically with $s$ while the 
right-hand-side decreases monotonically with $s$. Therefore, the 
intersection can be found with binary search. The value of $s$ at this 
intersection is then the optimal spacing for kernel $k$ given $m=2r+1$ 
discretization points according to our coverage criterion. Although the 
Fourier transforms of most stationary kernels are known analytically, we 
use the discrete FFT and numerical integration in our procedure to allow us 
to quickly adapt our method to new stationary kernels.

\subsection{Efficient Gradients Using the Permutohedral Lattice}

To perform maximum likelihood GP inference, we must optimize the 
marginal $\log$-likelihood given in \cref{eq:gp_mll} through gradient-based 
optimization. By relying on the BBMM method proposed in 
\citet{Gardner2018GPyTorchBM}, we need only a black-box function to compute 
kernel matrix MVMs and the derivative of the black-box function MVMs with 
respect to the kernel inputs. To remain computationally efficient, we must 
have efficient routines for \emph{both} the MVM and its derivative.
Here we show how to use the permutohedral lattice to also approximate the 
derivative of our MVM with respect to the input points.

Given a stationary kernel ${k: \reals^d \times \reals^d \to \reals}$,
vectors ${\{\mbf v_i\} \subseteq \reals^n}$ and 
${\{\xvec_i\} \subseteq \reals^d}$,  we have that,
\begin{align}
    \mbf u_i &:= \sum_j k(\mathbf{x}_i,\mathbf{x}_j)\mbf v_j = \sum_j k(d_{ij}^2)\mbf v_j ~,
\end{align}
where $d_{ij}^2=\norm{\mbf{x}_i-\mbf{x}_j}^2$.

The Jacobian vector products for this operation are: 
${\frac{\partial L}{\partial \mathbf{x}_n} = 
\sum_{i,j}\frac{\partial L}{\partial \mbf u_i}\frac{\partial k(d^2_{ij})}
{\partial \mathbf{x}_n}\mbf v_j}$. Noting that 
${\frac{\partial d^2_{ij}}{\partial \mathbf{x}_n} = 
2(\mathbf{x}_i-\mathbf{x}_j)(\delta_{in}-\delta_{jn})}$ where $\delta$ is
the Kronecker delta, we can rewrite the gradients as:
\begin{equation}
\frac{\partial L}{\partial \mathbf{x}_n} = 2\sum_{i,j}\frac{\partial L}{\partial \mbf u_i} k'(d^2_{ij})(\mathbf{x}_i-\mathbf{x}_j)(\delta_{in}-\delta_{jn})~,
\end{equation}
where $k'$ is the derivative with respect to $d^2$. As pointed out in 
\citet{Krhenbhl2011EfficientII}, even with the Gaussian filter $k'=k$, the 
computation does not seem to admit an implementation in terms of lattice 
filtering with the kernel. However, by splitting up the terms we can 
rewrite the sum as a lattice filtering of the gradients and outputs using 
the $k'$ kernel where either the input our output of the filter is 
elementwise multiplied by $\mathbf{x}$. Abbreviating 
${k_{ij}':=k'(d^2_{ij})}$ and 
${\mbf g_i := \frac{\partial L}{\partial \mbf u_i}}$, we have
\begin{equation}
\begin{split}
\frac{\partial L}{\partial \mathbf{x}_n} = 2\sum_{j}\big[\mbf v_nk_{nj}'\mathbf{x}_j \mbf g_j-v_n\mathbf{x}_nk_{nj}'\mbf g_j \\
+ \mbf{g}_nk_{nj}'\mathbf{x}_j\mbf v_j-\mbf g_n\mathbf{x}_nk_{nj}'\mbf v_j \big].    
\end{split}
\end{equation}

Using the automatic method for determining the filter stencil, we can then 
evaluate this expression using a single lattice filtering call with the 
kernel $k'$ on the input,
\begin{equation}
V = \mathrm{Concat}([\mathbf{x}\odot \mbf g,-\mbf g,\mathbf{x}\odot \mbf v, -\mbf v])~,
\end{equation}
where $\odot$ is elementwise multiplication. This way of rewriting the 
gradients allows us reuse our filtering implementation to compute the 
gradients of our MVMs efficiently.

\section{Experiments}

\begin{table*}[!t]
\caption{Standardized test root mean square error (RMSE) on various large-scale UCI regression datasets, averaged over 3 trials with two standard deviations. We use a $512$ inducing points for SGPR, and $100$ per dimension for SKIP. We were not able to run SKIP on our Titan RTX GPU (24GB) for the Houseelectric dataset, which we denote by Out of Memory (OOM). Numbers for Exact GP are taken from \citet{Wang2019ExactGP}. Best performing scalable GP methods are shown in \textbf{bold}.}
\label{table:rmse_lik}
\vskip 0.15in
\begin{adjustbox}{width=\textwidth}
\begin{sc}
\begin{tabular}{l|cccc|cccc}
\toprule
& \multicolumn{4}{c}{Test RMSE} & \multicolumn{3}{c}{Test NLL} \\
\midrule
Dataset & Exact GP & SGPR & SKIP & Simplex-GP & Exact GP & SGPR & SKIP & Simplex-GP \\
\midrule
Houseelectric & $0.054 \pm 0.000$ & $\mathbf{0.067} \pm 0.002$ & $\text{OOM}$ & $0.079 \pm 0.002$ & $-0.207 \pm 0.001$ & $\mathbf{-1.242} \pm 0.057$  & $\text{OOM}$ & $0.756 \pm 0.075$ \\
Precipitation & $0.937 \pm 0.000$ & $1.033 \pm 0.004$ & $1.032 \pm 0.001$ & $\mathbf{0.939} \pm 0.001$ & $\mathrm{-}$ & $1.437 \pm 0.006$ & $1.451 \pm 0.001$ & $\mathbf{1.397} \pm 0.001$ \\
Keggdirected & $0.083 \pm 0.001$ & $0.380 \pm 0.018$ & $0.487 \pm 0.005$ & $\mathbf{0.095} \pm 0.002$ & $-0.838 \pm 0.031$ & $0.985 \pm 0.007$ & $0.996 \pm 0.013$  & $\mathbf{0.797} \pm 0.031$ \\
Protein & $0.511 \pm 0.009$ & $0.579 \pm 0.003$ & $0.817 \pm 0.012$ & $\mathbf{0.571} \pm 0.003$ & $0.960 \pm 0.003$ & $\mathbf{0.982} \pm 0.059$ & $1.213 \pm 0.020$ & $1.406 \pm 0.048$ \\
Elevators & $0.399 \pm 0.011$ & $\mathbf{0.356} \pm 0.006$ & $0.447 \pm 0.037$ & $0.510 \pm 0.018$ & $0.626 \pm 0.043$ & $1.031 \pm 0.230$ & $\mathbf{0.869} \pm 0.074$  & $1.600 \pm 0.020$ \\
\bottomrule
\end{tabular}
\end{sc}
\end{adjustbox}
\end{table*}

In our experiments, we establish that (i) our proposed discretization 
scheme to approximate a continuous stationary kernel has low  
error, (ii) leveraging the permutohedral lattice allows us to operate on 
large scale datasets with over a million data points without demanding 
significant runtime and memory requirements unlike other methods, and (iii) our 
method closes the performance gap of SKI methods relative to exact Gaussian 
processes at a significantly lower computational cost. All code
to reproduce results is available at \codeurl. All
hyper-parameters are documented in \cref{sec:hypers}.

We note that comparisons to KISS-GP are not possible with our evaluation 
datasets as due to KISS-GP's exponential scaling with dimension. Within the 
SKI framework, however, our method Simplex-GP provides a scalable 
alternative for KISS-GP to dimensions much larger than originally possible, 
and outperform SKIP in test \emph{root mean squared error} by a significant 
margin. Furthermore, on large scale datasets ($n\sim 10^6$), we observe that MVMs using 
the permutohedral lattice are 10x faster than exact MVMs computed with the
highly efficient KeOps library \citep{charlier2020kernel}, and the asymptotics suggest
even greater gains on yet larger datasets.

\subsection{Evaluating Discretization Error}

\begin{figure}[!ht]
\centering
\includegraphics[width=\linewidth]{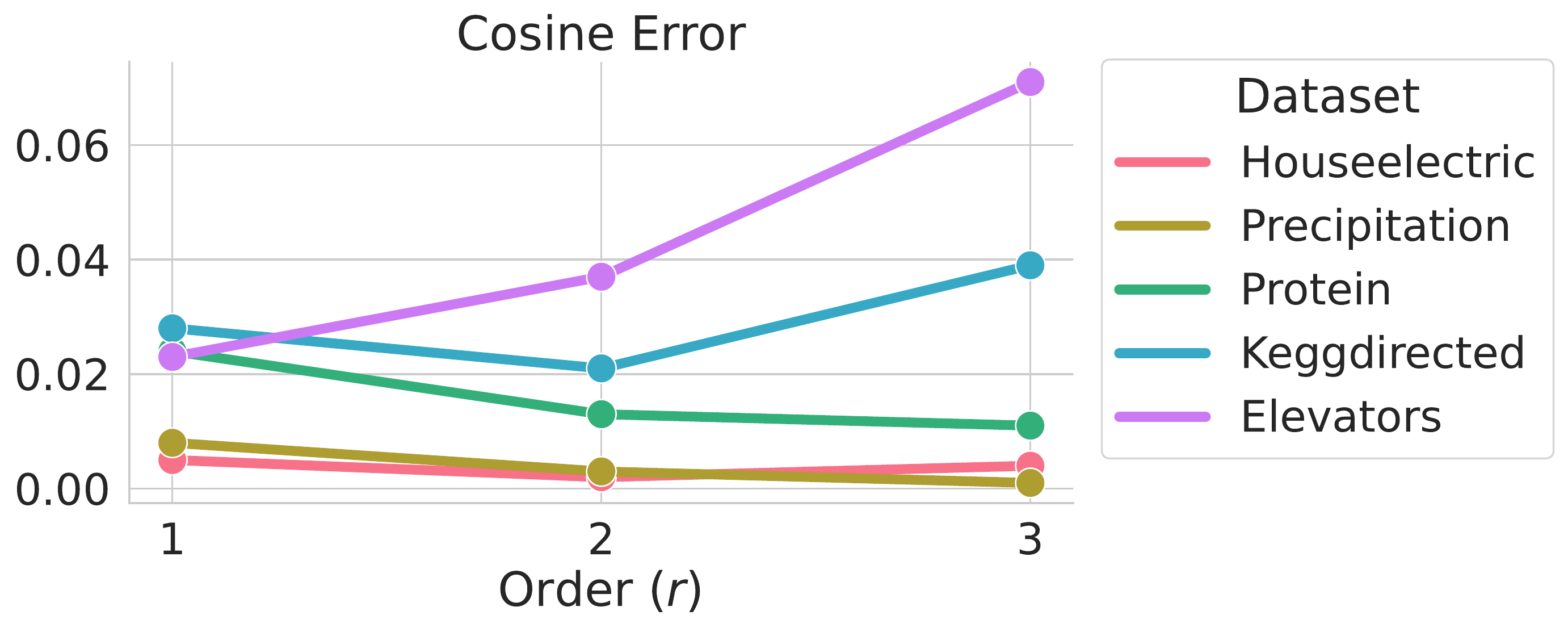}
\caption{
For different sizes of the blur stencil denoted by order $r$, we show the 
MVM error for the resultant vector $\hat{\mbf{z}}$ with Simplex-GPs, as compared 
to $\mbf{z}$ computed with exact GPs using KeOps. We compute the cosine error 
between the vectors as $1 - \frac{\langle \mbf{z},\hat{\mbf{z}} \rangle}
{\norm{\mbf{z}}\norm{\hat{\mbf{z}}}}$. Low values indicate that the two vectors 
are closely aligned. Importantly, increasing the order of the blur stencil does 
not always reduce the error, because the truncation in the blur step affects the 
approximation.}
\label{fig:mvm_err}
\end{figure}

The fundamental unit of computation in Simplex-GP is a 
\emph{matrix-vector multiplication} (MVM). It is not feasible to store the 
complete covariance matrix for many of the large scale datasets we 
consider. MVMs, however, only require the storage of the resultant vector, 
and therefore has nimble memory requirements. Consequently, instead of 
comparing dense covariance matrices between exact GPs and Simplex-GPs on 
the benchmark datasets, we compute the cosine error of the resultant vector 
with respect to results from exact GPs via KeOps 
\citep{charlier2020kernel}. \cref{fig:mvm_err} shows the errors achieved 
for different sizes of the blur stencil, denoted by order $r$ in 
\cref{sec:disc_kernels}.

\subsection{Computational Gains from Lattice Sparsity}

\begin{figure}[!ht]
\centering
\includegraphics[width=\linewidth]{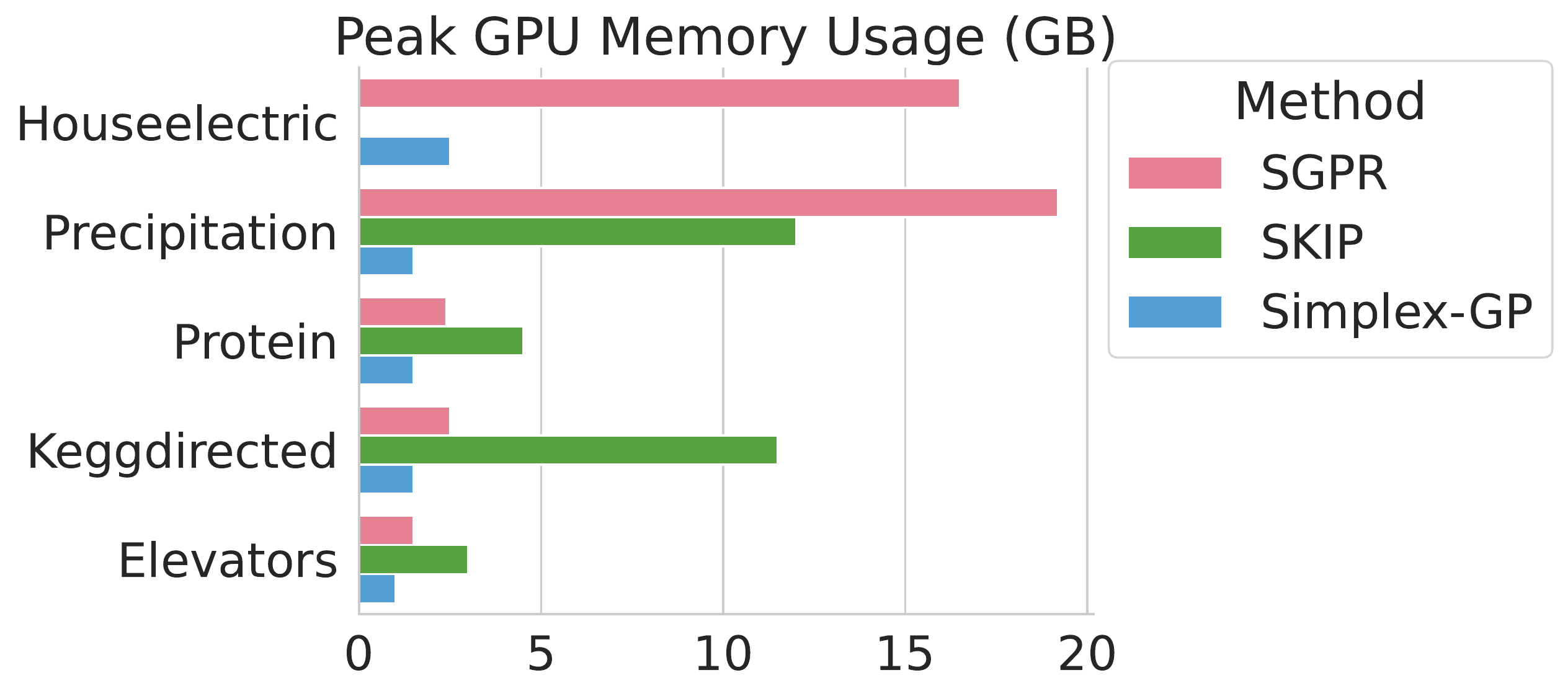}
\caption{
Owing to the significant gains in the number of lattice points generate as 
reported by \cref{table:n_lattice}, we see that the approximate peak GPU 
memory usage of our proposed approximation is significantly lower, allowing 
operations on large-scale datasets. We were unable to fit the Houselectric 
dataset with SKIP on one Titan RTX GPU (24GB).}
\label{fig:mem_usage}
\end{figure}

One of the significant advantages of using the permutohedral lattice 
instead of the Caretesian grid to embed our training inputs is that we can 
leverage the gains from the efficient packing afforded by the lattice. The 
permutohedral lattice provably provides the best covering density up to 
dimension $5$, and is the best known lattice packing up to dimension $20$ 
\citep{Conway1988SpherePL,Baek2012LatticeBasedHG}.

\begin{table}[!ht]
\caption{
Embedding a set of $n$ $d$-dimensional inputs onto a \emph{permutohedral 
lattice} generates $m$ lattice points with 
${m \leq L := n \times (d + 1)}$ in the worst case. We quantify the 
sparsity for various UCI regression benchmark datasets as the ratio of 
lattice points generated to the maximum possible, $m / L$. We observe a 
significantly fewer number of lattice points generated for our benchmark 
datasets.}
\label{table:n_lattice}
\vskip 0.15in
\begin{adjustbox}{width=\linewidth}
\begin{sc}
\begin{tabular}{lcccc}
\toprule
Dataset & $n$ & $d$ & $m$ & $m/L$ \\
\midrule
Houseelectric & 2,049,280 & 11 & 1,000,190 & $0.04$ \\
Precipitation & 628,474 & 3 & 480 & $0.003$ \\
Keggdirected & 48,827 & 20 & 122,755 & $0.12$ \\
Protein & 45,730 & 9 & 14,715 & $0.03$ \\
Elevators & 16,599 & 17 & 204,761 & $0.69$ \\
\bottomrule
\end{tabular}
\end{sc}
\end{adjustbox}
\end{table}

As noted earlier, each input embedded into the permutohedral lattice can 
generate at most $d + 1$ neighbors. We insert each generated lattice point 
into a hashtable. \cref{table:n_lattice} shows the total number of such 
lattice points generated on each of standardized benchmark datasets, and 
compute the ratio to the worst-case scenario, such that a ratio of $1$ 
corresponds to having generated $L = n \times (d + 1)$ lattice points. By 
storing and manipulating only a sparse lattice, instead of one where all 
the interior points in the convex hull of the data are generated, the gains 
from the sparse structure are actually much more than just $m/L$, and 
exponentially better both in the dimension and the chosen lengthscale.

\begin{figure}[!ht]
\centering
\includegraphics[width=\linewidth]{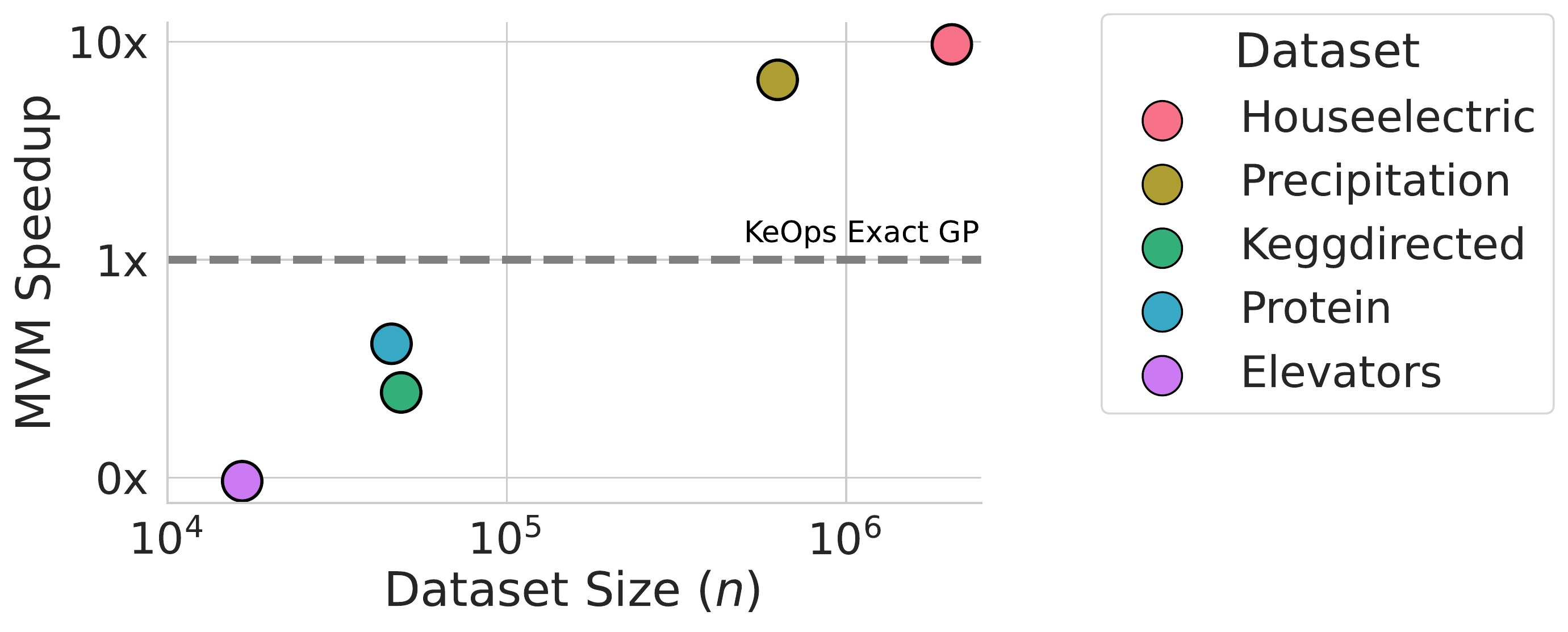}
\caption{We compare the speed of our Simplex-GP MVMs at order $r=1$ on all benchmark datasets against exact MVMs with KeOps \citep{charlier2020kernel}, both using one Titan RTX GPU. We achieve up to 10x speedups for large datasets with over $10^5$ data points.}
\label{fig:mvm_speedup}
\end{figure}

We observe that the ratios for all datasets are significantly lower than 
$1$. A ratio closer to one indicates that the input dataset has extremely 
high variance. Learning in such a scenario would be significantly 
challenging for any method. As a consequence, we are able to keep the 
memory usage considerably low, as reported in \cref{fig:mem_usage}. 
\cref{fig:mvm_speedup} shows significant speedups for large scale datasets.

\subsection{Performance on Benchmark Datasets}

We focus on Gaussian process regression, and work with datasets from the 
UCI repository \citep{Dua2019}, which either have large number of training 
samples, or have a large dimension, to demonstrate the advantages of 
Simplex-GP.

The data is randomly split into a $4/9\mathrm{-}2/9\mathrm{-}3/9$ 
train-validation-test split. All datasets are standardized using the 
training data to have zero mean and unit variance. We report the 
standardized \emph{root mean squared error} (RMSE) and the marginal 
$\log$-likelihood over the test set, with a standard deviation over 3 
independent runs. All methods are trained using a learning rate of 0.1 with 
the Adam optimizer in GPyTorch \citep{Gardner2018GPyTorchBM}. As noted in 
\citet{Gardner2018GPyTorchBM}, we set the CG error tolerance to $1.0$ 
during training, but use a lower CG tolerance of $0.01$ during evaluation, 
which does not significantly hurt performance.

For SGPR \citep{Titsias2009VariationalLO}, we use the typical value of 
$m = 512$ inducing points, as reported in literature. For SKIP 
\citep{Gardner2018ProductKI}, we use a total of $m = 100$ inducing points 
per dimension, which is significantly larger than originally reported. We 
were, however, not able to fit inference with SKIP on a single GPU device, 
even for small values of $m$.

\cref{table:rmse_lik} reports the standardized test RMSE and the 
corresponding negative $\log$-likelihoods for our benchmark datasets from 
the UCI repository. We are able to significantly outperform SKIP 
\citep{Gardner2018ProductKI}, which is one of the more scalable methods 
within the SKI framework \citep{Wilson2015ThoughtsOM}. Further, on most 
datasets, we are able to match or outperform the approximate inducing point 
method SGPR \citep{Titsias2009VariationalLO}. 

Finally, in \cref{sec:learned_ls}, we compare the lengthscales learned by
Simplex-GPs to those learned by exact GPs (with KeOps) for all our benchmark
datasets, noting that the performance of Simplex-GPs is not simply a
coincidental artifact of optimization, but provides qualitatively similar
automatic relevance determination (ARD).

\subsection{Practical Considerations}
\label{sec:practical}


The Simplex-GP makes use of iterative methods such as conjugate gradients (CG). Conjugate gradients are highly effective for scalable Gaussian processes \citep{Gardner2018GPyTorchBM}, but are also sensitive to design decisions and prone to numerical instability. Here we provide guidance to help practitioners seamlessly deploy the Simplex-GP in their own applications. We note many of these considerations are not specific to the Simplex-GP and have broad relevance to CG based GPs.

Conjugate gradients (CG) enable the solution to a rank-$n$ linear system in $p \ll n$ iterations up to machine precision (and exact when $p = n$). In 
practice, we run CG up to a prescribed error tolerance, or some maximum 
prescribed value of $p$, whichever is achieved earlier. Using low 
tolerance or high $p$ can add a significant runtime cost. \citet{Wang2019ExactGP} 
recommend that an error tolerance of $10^{-2}$ provides a reasonable trade-off 
between training time and performance, although this recommendation mostly holds for good RMSE performance rather than good test likelihood.
Moreover, this relatively high tolerance can lead to unstable 
training such that the MLL may not always improve monotonically.
We illustrate this pathology in \cref{fig:unstable_train} (\cref{sec:unstable_train}). 
In addition, early truncations due to a small value of $p$ can introduce bias 
\citep{andres2021rrcg}. To avoid adverse impact of these pathologies, we rely on early 
stopping, i.e. we use the RMSE on a held-out validation set to select the best 
model as a result of the MLL optimization.

\begin{table}[!ht]
\caption{
We compare training runtime on a single epoch of Simplex-GPs with 
different error tolerance values for conjugate gradients, and observe that 
the Russian Roulette estimator for randomized CG truncations (RR-CG) 
\citep{andres2021rrcg} is able to sustain a much better runtime. This 
allows us to stabilize training, without compromising the speedup gains 
from Simplex-GPs by a significant margin. The approximate runtime ranges 
below represent a the time taken by a single epoch during training in 
\textbf{seconds}.}
\label{table:cg_runtimes}
\vskip 0.15in
\begin{sc}
\begin{adjustbox}{width=\linewidth}
\begin{tabular}{lccc}
\toprule
Dataset & CG $(10^{-2})$ & CG $(10^{-4})$ & RR-CG $(10^{-8})$ \\
\midrule
Houseelectric & 450-600 & 2000-4000 & 950-1200  \\
Precipitation & 15-20  & 20-40 & 30-40  \\
Keggdirected & 40-50 & 320-340 & 75-100   \\
Protein & 8-10 & 40-60 & 15-20 \\
Elevators & 18-20 & 25-30  & 20-25 \\
\bottomrule
\end{tabular}
\end{adjustbox}
\end{sc}
\end{table}

For Simplex-GPs, we find that while these numerical instabilities can be 
alleviated by reducing the CG tolerance to $10^{-4}$ (see 
\cref{fig:unstable_train}, \cref{sec:unstable_train}), there is a significant training slowdown (\cref{table:cg_runtimes}). In recent work \citet{andres2021rrcg} 
propose randomized truncations for bias-free conjugate gradients, termed RR-CG, to remedy these challenges with using CG based inference. In \cref{table:cg_runtimes}, we find that RR-CG is useful in avoiding the pathologies while maintaining an acceptable runtime.

The computational gains from
Simplex-GPs are a consequence of exploiting the geometry of the data. A proxy for the potential gains in the MVMs is the sparsity ratio of the lattice as 
in \cref{table:n_lattice}. 
Qualitatively, we expect especially large acceleration in datasets where data overlap or large lengthscales lead to a smaller fraction of inducing points $m/L$ over the maximum possible. In general, we expect Simplex-GPs to be especially valuable on large training sets (more than $10^5$ points), with moderate input dimensionality (between about $3$ and $20$ dimensional inputs).

\section{Conclusion}

Our work demonstrates the benefits of cross-pollinating concepts in image 
filtering with Gaussian process inference. By leveraging the 
permutohedral lattice for accelerated filtering, we are able to alleviate the curse of dimensionality in \emph{Structured Kernel Interpolation}.
While Simplex-GP results in an asymptotic speedup over standard inference,
the runtime constants are large, meaning that this speedup is mostly realized for large datasets.
We hope that the promising results here can spur on future work, helping to further broaden the applicability of Simplex-GPs, to higher dimensional problems and a greater range of datasets.

\section*{Acknowledgements}

We would like to thank Wesley J. Maddox for insightful discussions. 
This research is supported by an Amazon Research Award, NSF I-DISRE 193471, NIH R01DA048764-01A1, NSF IIS-1910266, and NSF 1922658 NRT-HDR: FUTURE Foundations, Translation, and Responsibility for Data Science.

\bibliography{references}
\bibliographystyle{icml2021}

\clearpage
\appendix

\twocolumn[
\icmltitle{Appendix for \\ SKIing on Simplices: Kernel Interpolation on the Permutohedral Lattice for Scalable Gaussian Processes}

\icmlsetsymbol{equal}{*}

\begin{icmlauthorlist}
\icmlauthor{Sanyam Kapoor}{equal,nyu}
\icmlauthor{Marc Finzi}{equal,nyu}
\icmlauthor{Ke Alexander Wang}{stanford}
\icmlauthor{Andrew Gordon Wilson}{nyu}
\end{icmlauthorlist}

\icmlaffiliation{nyu}{New York University, NY, USA}
\icmlaffiliation{stanford}{Stanford University, CA, USA}

\icmlcorrespondingauthor{Sanyam Kapoor}{sanyam@nyu.edu}

\icmlkeywords{Gaussian processes, permutohedral lattice}

\vskip 0.3in
]


\section{Hyperparameters}
\label{sec:hypers}

\cref{table:sgp_hypers} documents all the hyperparameters used for training
Simplex-GPs. All kernels use automatic relevance determination (ARD).
We find that higher values of $r$ (e.g. $2$ or $3$) do not meaningfully improve 
the test RMSE performance, but significantly increase the training time.

\begin{table}[!ht]
\caption{
We document all the settings and hyperparameters involved in training
Simplex-GPs.
}
\label{table:sgp_hypers}
\vskip 0.15in
\begin{sc}
\begin{adjustbox}{width=\linewidth}
\begin{tabular}{l|c}
\toprule
Hyperparameter & Value(s) \\
\midrule
Max. Epochs & $100$  \\
Optimizer & $\mathrm{Adam}$ \\
Learning Rate & $0.1$ \\
CG Train Tolerance & $1.0$ \\
CG Eval/Test Tolerance & $0.01$ \\
Max. CG Iterations & $500$ \\
CG Pre-conditioner Rank & $100$ \\
Max. Lanczos Iterations & $100$ \\
Kernel Family & \{ $\mathrm{Mat\acute{e}rn}$-3/2, $\mathrm{RBF}$ \} \\
Blur Stencil Order ($r$) & 1 \\
Min. Likelihood Noise ($\sigma^2$) & \{ $10^{-4}$, $10^{-1}$ \} \\
\bottomrule
\end{tabular}
\end{adjustbox}
\end{sc}	
\end{table}

\section{Visualizing Training Instabilities}
\label{sec:unstable_train}

\begin{figure*}[!ht]
\centering
\begin{tabular}{cc}
\includegraphics[width=.48\linewidth]{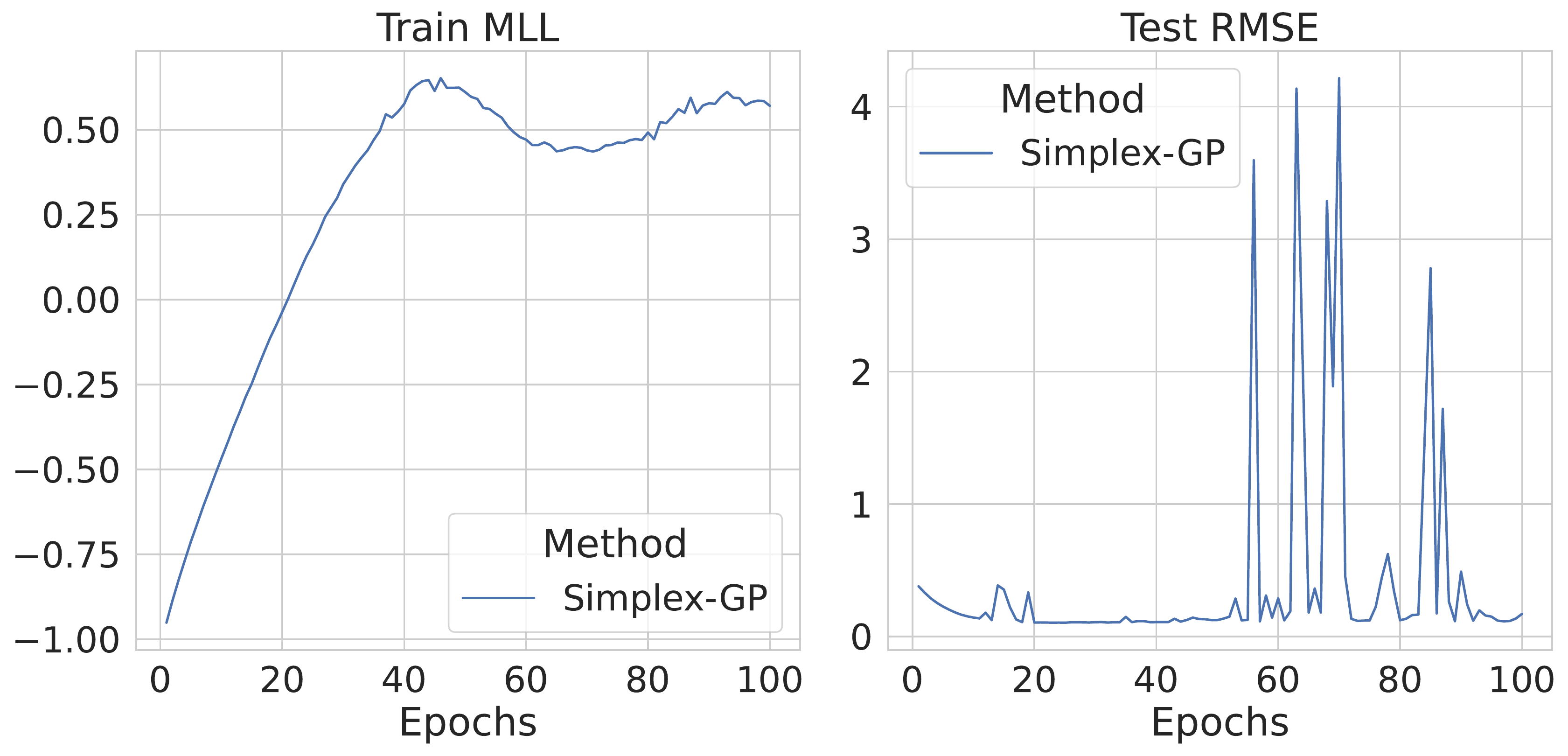} &
\includegraphics[width=.48\linewidth]{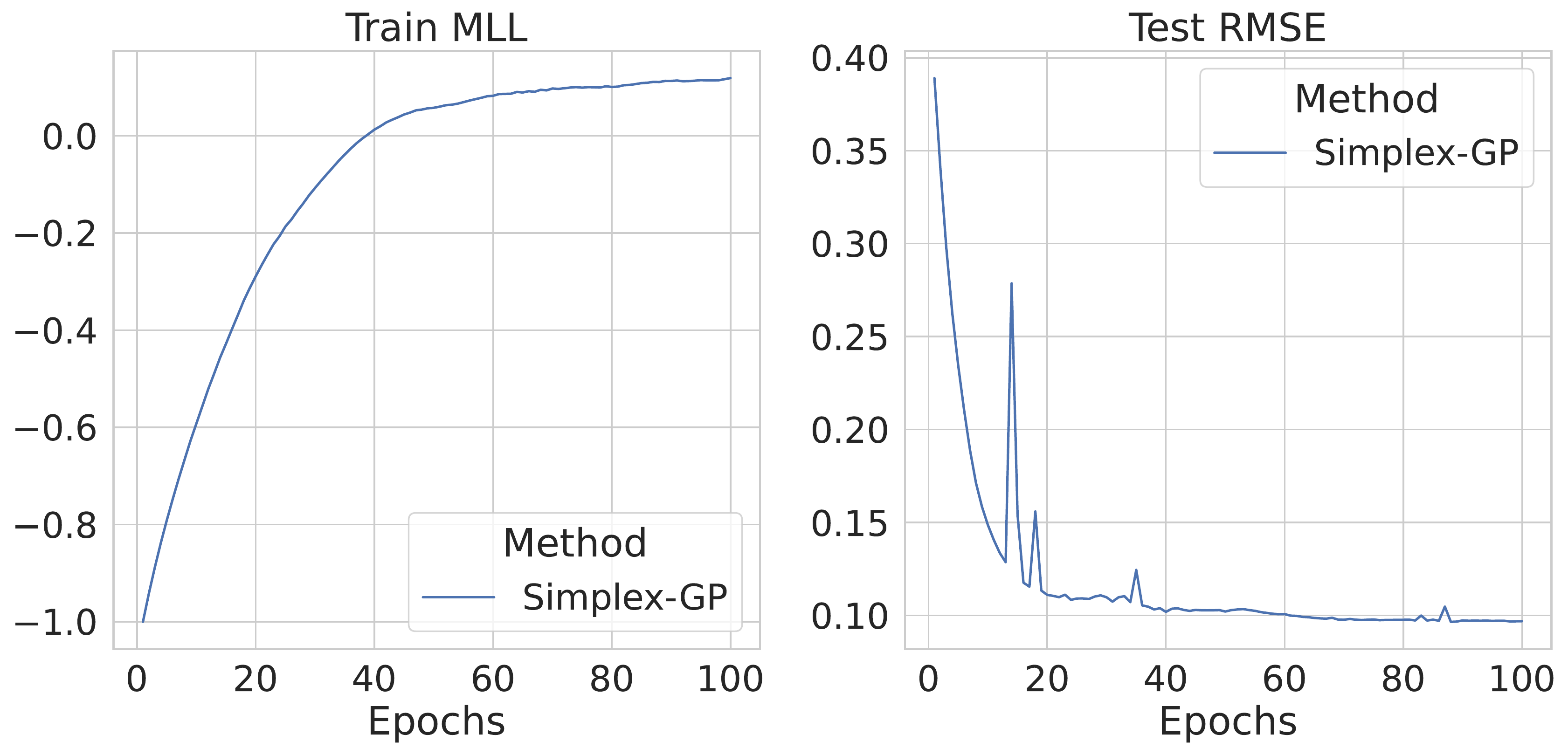} \\
(a) & (b)
\end{tabular}
\caption{
We visualize the pathology discussed in \cref{sec:practical}, when using 
conjugate gradients (CG) on the \texttt{keggdirected} dataset. We observe similar
behavior for other datasets too. (a) Using a high CG error tolerance of $1.0$ during 
training leads to non-monotonic improvements in the train marginal $\log$-likelihood 
(MLL) due to convergence issues in CG. More significantly, this makes the test RMSE 
curves look unstable. (b) By simply reducing the CG error tolerance to $10^{-4}$, we are 
able to stabilize these curves, behaving more favorably.
}
\label{fig:unstable_train}
\end{figure*}

We visualize the training instabilities that arise as a consequence of using a 
high CG tolerance value. As noted in \cref{sec:practical}, we follow the 
recommendation of \citet{Wang2019ExactGP}, and use a CG tolerance of $1.0$ 
during training and $0.01$ during validation and test. We find that the train 
MLL does not improve monotonically, due to lack of CG convergence, often
owing to early truncation. This leads to undesirable behavior in the test RMSE
as visualized in \cref{fig:unstable_train}(a).

As addressed in \cref{sec:practical}, a more stable training run is
achieved by simply reducing the tolerance to $10^{-4}$, as visualized in
\cref{fig:unstable_train}(b). But this leads to a significant slowdown,
defeating the computational gains from Simplex-GPs. Therefore, this remains
a noteworthy design decision for practical usage.

\section{Comparing Learned Lengthscales with Exact GPs}
\label{sec:learned_ls}

\begin{figure*}[!ht]
\centering
\begin{tabular}{ccc}
\includegraphics[width=.31\linewidth]{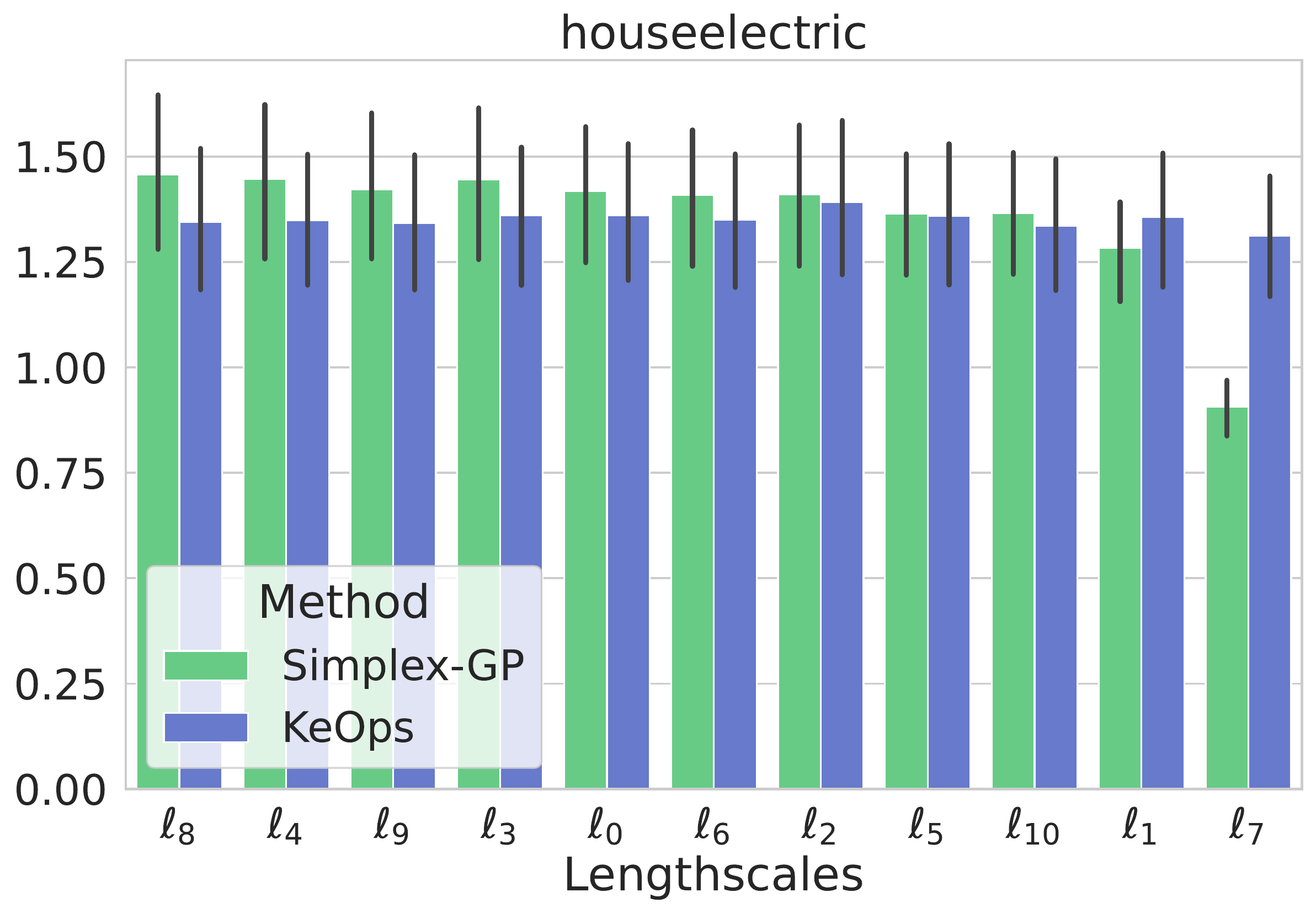} & 	\includegraphics[width=.31\linewidth]{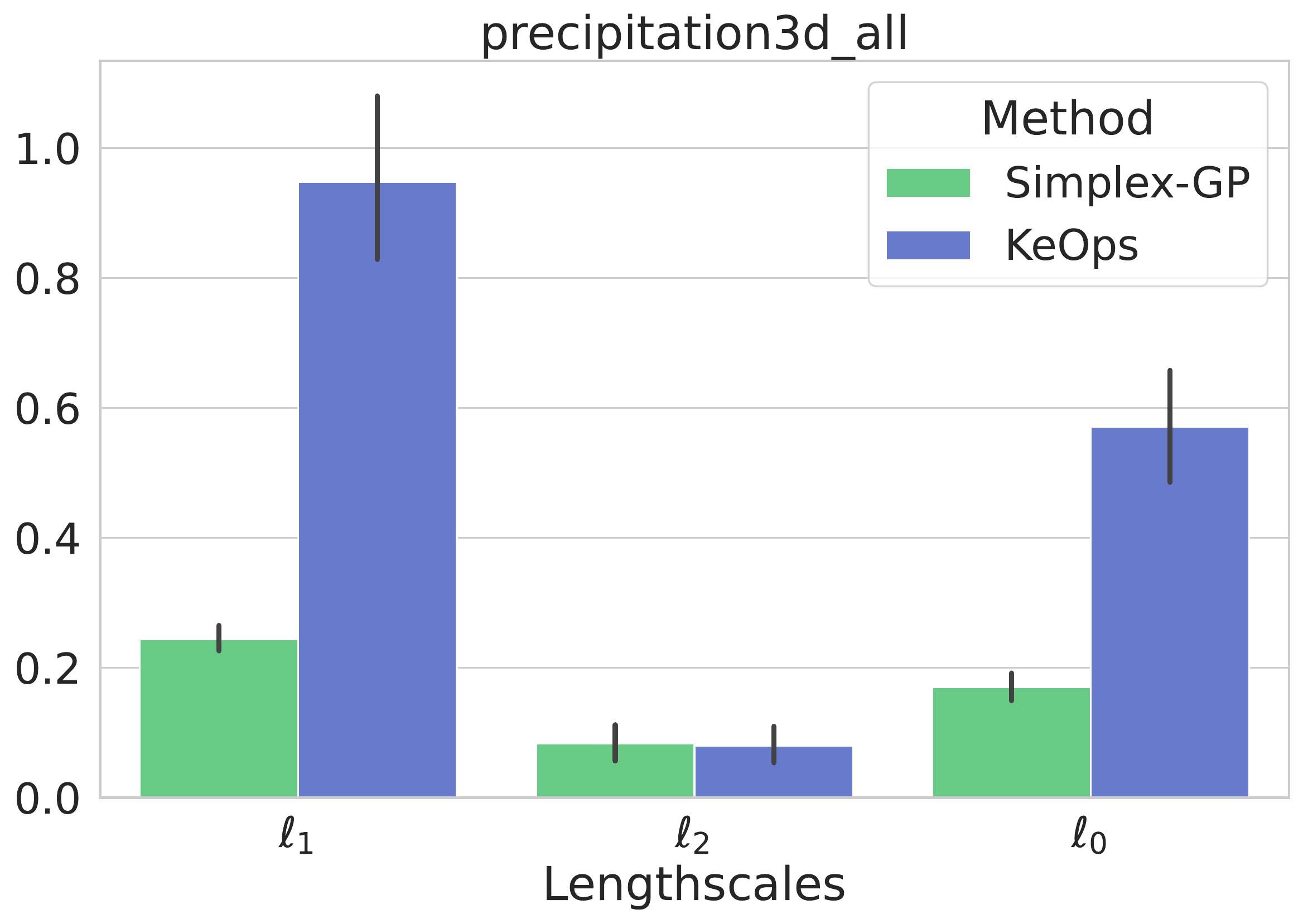} &
\includegraphics[width=.31\linewidth]{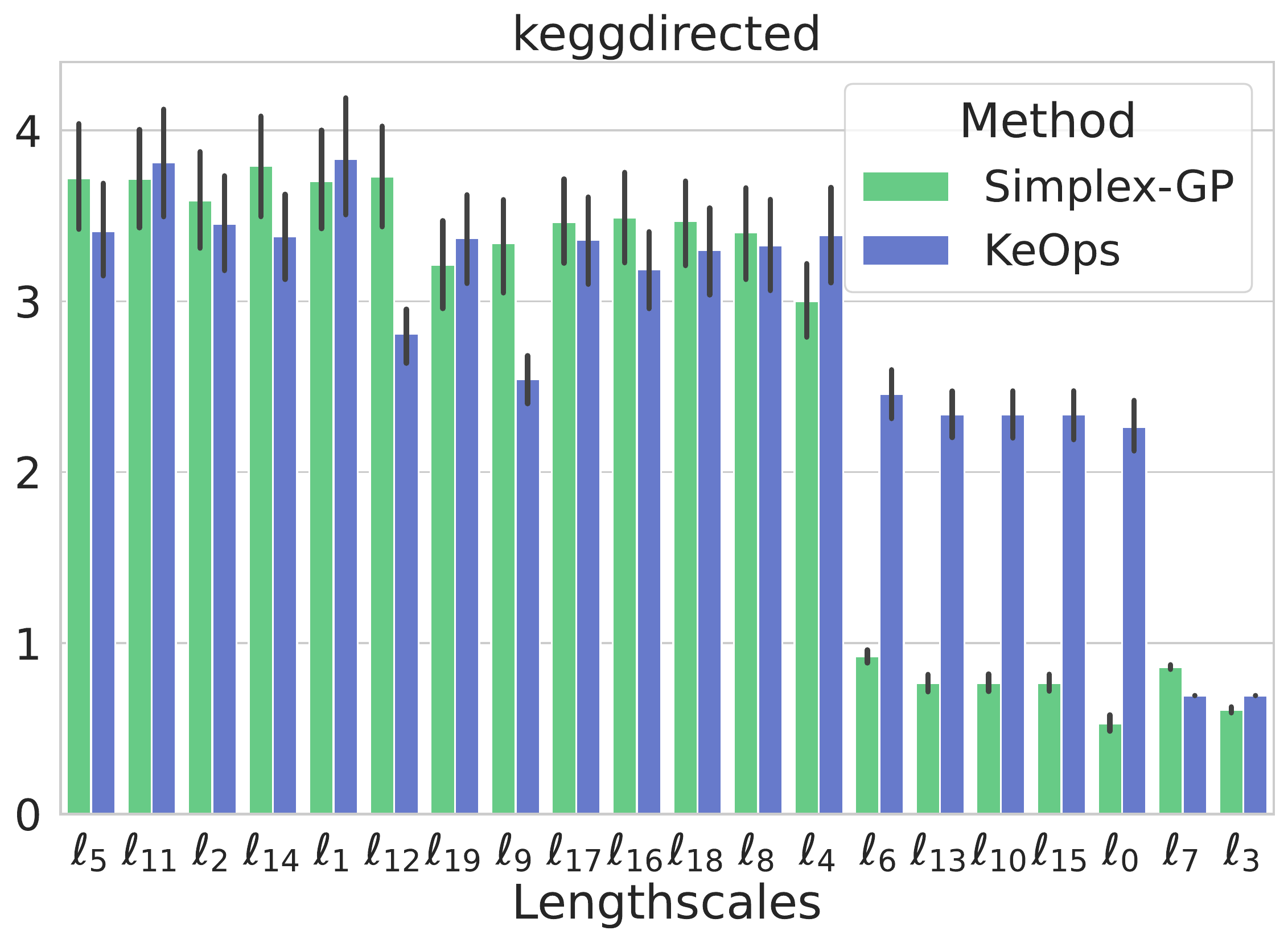} \\ \includegraphics[width=.31\linewidth]{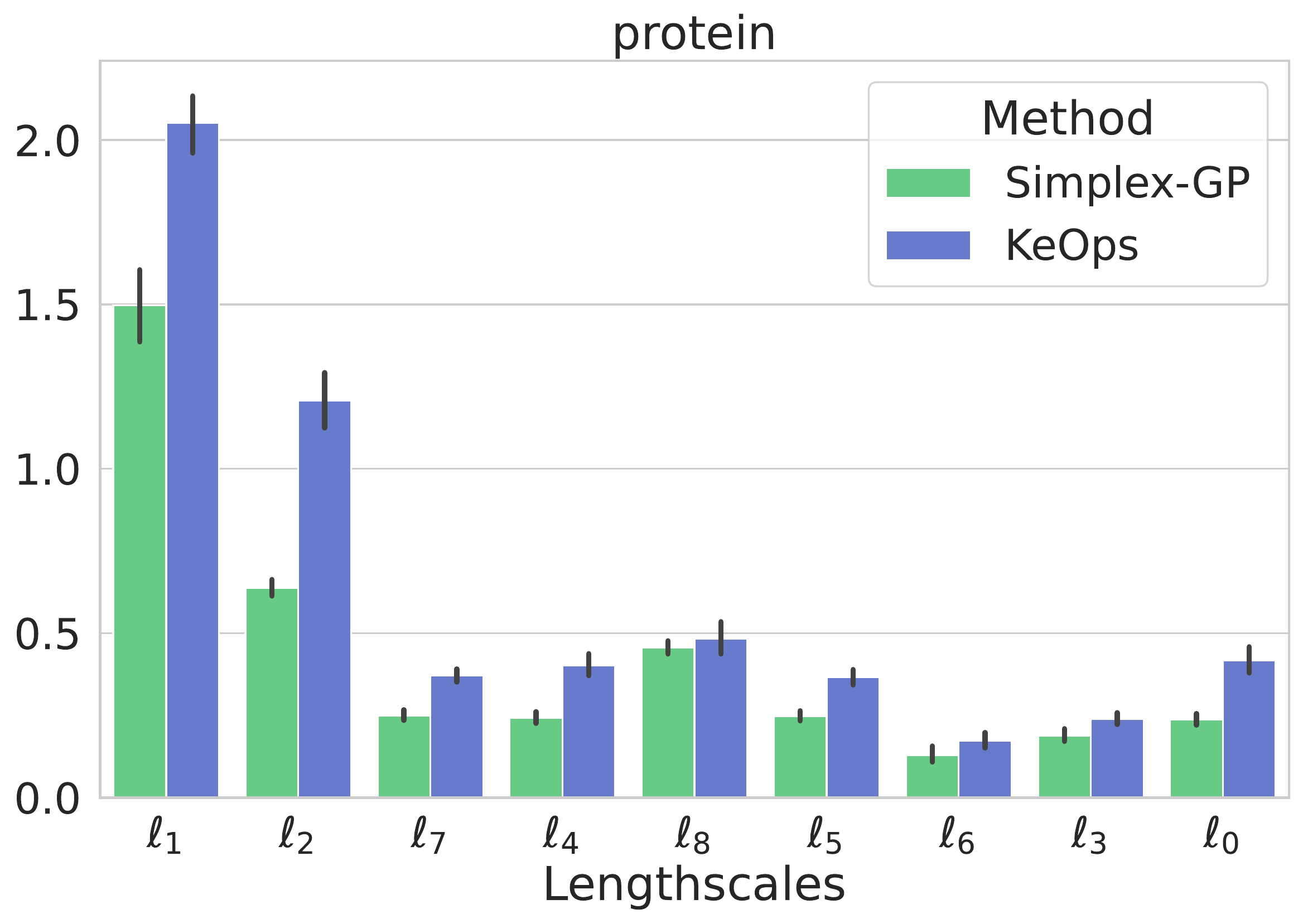} &
\includegraphics[width=.31\linewidth]{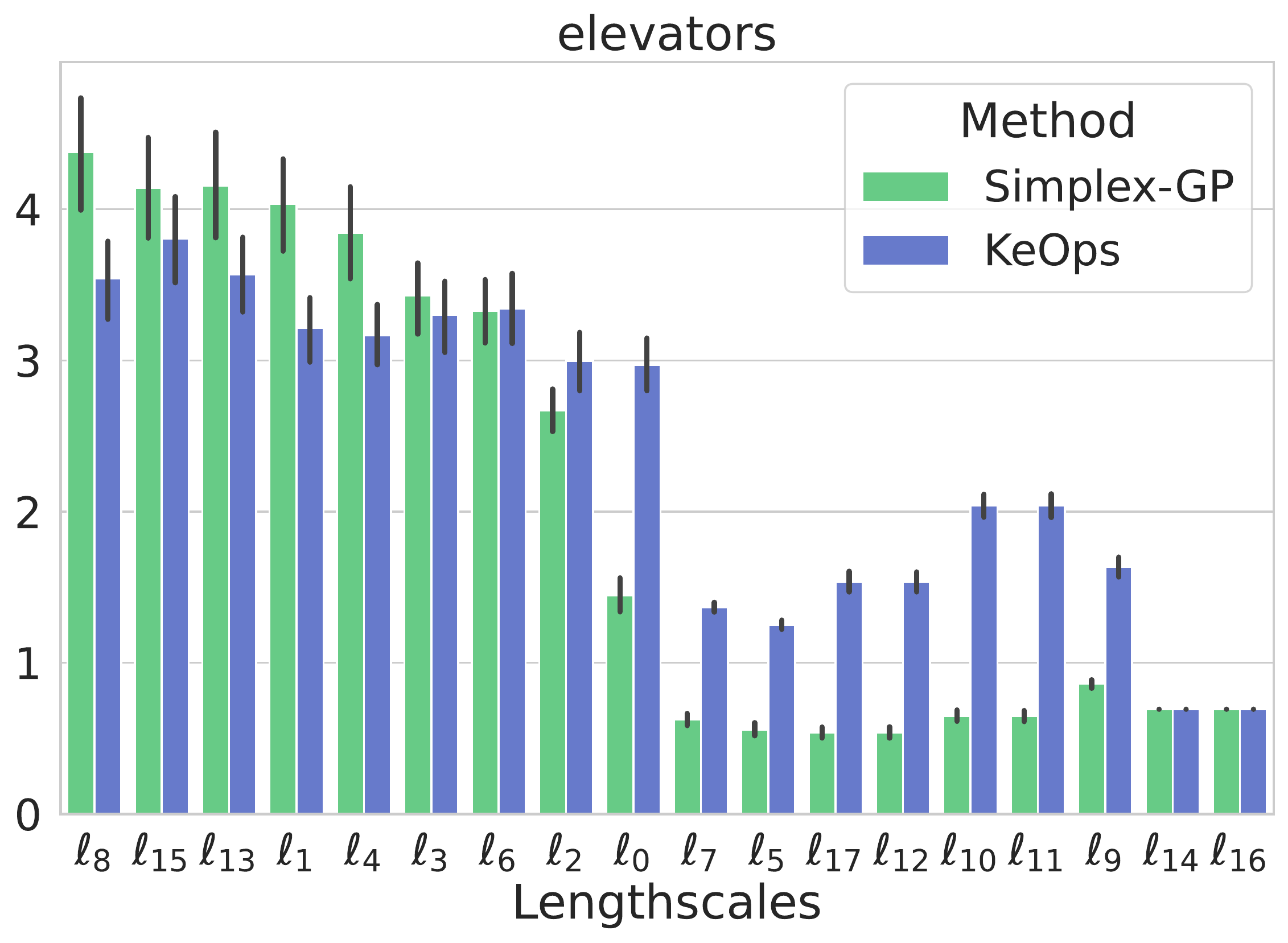} & 
\end{tabular}
\caption{
For all our benchmark UCI datasets, when comparing the lengthscales between those 
learned by Simplex-GPs, and those learned by exact GPs using KeOps, we find that 
the learned values agree in terms of determined relevance. The label $\ell_d$ 
refers to the lengthscale learned for dimension $d$.
}
\label{fig:learned_ls}
\end{figure*}

When comparing the results from the Simplex-GP approximation to exact GPs via 
KeOps \citep{charlier2020kernel}, we find that the learned lengthscales for the 
$\mathrm{Mat\acute{e}rn}$-3/2 ARD kernel agree qualitatively, i.e. the relevance 
determined by Simplex-GPs corresponds to the relevance determined by KeOps too. 
In many cases, these agree quantitatively too. This is visualized in 
\cref{fig:learned_ls}. The learned scale factors for the kernels are often 
different, partially accounting for the difference in the magnitude of
lengthscales.

This hints that the approximations constructed by Simplex-GPs are meaningful in 
practice, and similar in quality to exact GPs, than the performance just being
coincidental artifact of optimization.

\end{document}